\pdfoutput=1

\documentclass[11pt]{article}

\usepackage{EMNLP2023}

\usepackage{times}
\usepackage{latexsym}
\usepackage{graphicx}
\usepackage{multirow}
\usepackage{color}
\usepackage{booktabs}
\usepackage[T1]{fontenc}

\usepackage[utf8]{inputenc}

\usepackage{microtype}

\usepackage{inconsolata}

\title{Make a Choice! Knowledge Base Question Answering with In-Context Learning}

\author{Chuanyuan Tan$^1$, Yuehe Chen$^1$, Wenbiao Shao$^1$, Wenliang Chen$^1$ \\ \textbf{Zhefeng Wang}$^2$, \textbf{Baoxing Huai}$^2$ \and \textbf{Min Zhang}$^1$\\
$^1$Institute of Artificial Intelligence, School of Computer Science and Technology, \\Soochow University, China \\ $^2$Huawei Cloud, China \\
\texttt{\{cytan17726, yhchen2020, wbshao\}@stu.suda.edu.cn}\\
\texttt{\{wlchen, minzhang\}@suda.edu.cn}, \texttt{\{wangzhefeng, huaibaoxing\}@huawei.com}
}

\begin{document}
\maketitle
\begin{abstract}

Question answering over knowledge bases (KBQA) aims to answer factoid questions with a given knowledge base (KB). Due to the large scale of KB, annotated data is impossible to cover all fact schemas in KB, which poses a challenge to the generalization ability of methods that require a sufficient amount of annotated data.
Recently, LLMs have shown strong few-shot performance in many NLP tasks. We expect LLM can help existing methods improve their generalization ability, especially in low-resource situations.
In this paper, we present McL-KBQA, a framework that incorporates the few-shot ability of LLM into the KBQA method via ICL-based multiple choice and then improves the effectiveness of the QA tasks. Experimental results on two KBQA datasets demonstrate the competitive performance of McL-KBQA with strong improvements in generalization. We expect to explore a new way to QA tasks from KBQA in conjunction with LLM, how to generate answers normatively and correctly with strong generalization.
\end{abstract}

\section{Introduction}
\begin{figure*}[t]
    \centering
    \includegraphics[width=16cm]{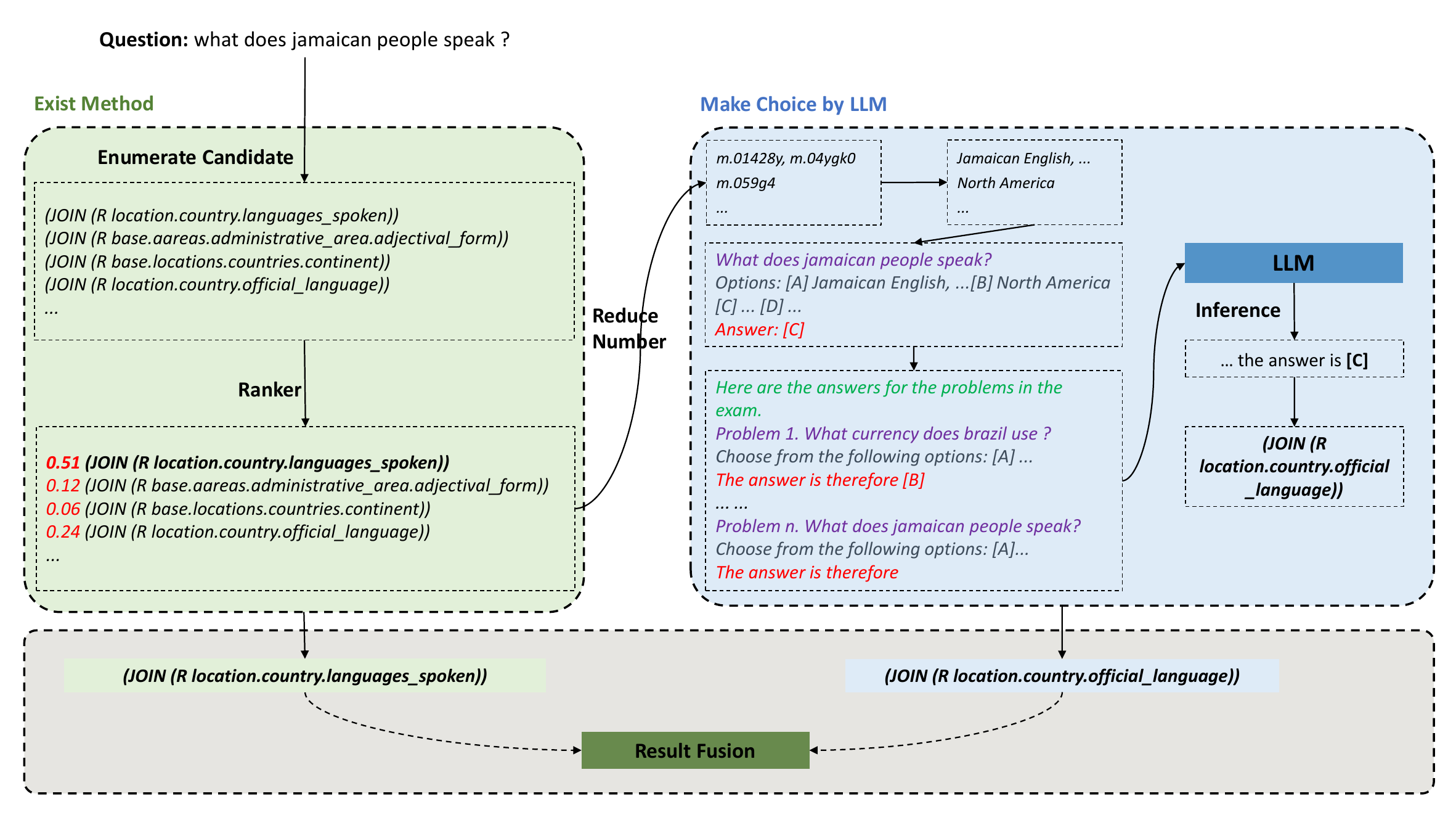}
    \caption{Overview of McL-KBQA framework. Given a question, we use a rank-based method to enumerate and score logical form candidates. With candidates provided by rank-based method, we transform original questions into multiple-choice form and construct prompts for LLM to make choices via ICL. We fuse result between existing method and LLM to determine the final logical form. The final logical form is executed over the KB to yield the answer.}

    \label{fig:method}
\end{figure*}

Knowledge Base Question Answering (KBQA) is a task to answer natural language questions over KBs. Many existing methods have achieved good performances\citep{chen-etal-2021-retrack,ye2021rng,gu-su-2022-arcaneqa} in this task. However, due to the large scale of the KB, the existing annotated training data can only cover a small portion of the information region of the KB. Although existing methods try to extend the QA model to a larger perceptual range as much as possible, the long-tail information in the KG, which is difficult for existing methods to generalize the semantics of the questions, is still beyond the ability of existing methods. Long-tail information necessitates that the KBQA methods have stronger generalization abilities, which can be viewed as a low-resource environment for few-shot learning. 

Recently, large language models (LLMs) have shown strong comprehension and reasoning ability in many tasks via few-shot in-context learning (ICL) \citep{brown2020language,lewkowycz2022solving,ma2023large}. We look forward to its effectiveness on KBQA.

Therefore, how to apply LLMs to KBQA tasks is a topic worth discussing. \citep{omar2023chatgpt} and \citep{tan2023evaluation} evaluate ChatGPT as a KBQA system by providing questions directly to LLMs. One challenge is that LLM outputs generative long text which is different from the gold answers of KBQA, consisting of phrases corresponding to entities in KB. Due to this difference, it is necessary to manually evaluate or adopt a strategy of matching answer phrases from the generative long text, which is costly or cumbersome to evaluate. Additionally, LLM fails to answer factual questions sometimes, which is called Hallucination \citep{ji2023survey}. How to use the knowledge in KB to alleviate hallucination of LLM is another challenge.\par

To address the challenges above, we propose McL-KBQA, a framework to solve KBQA problems in the form of making choices. Based on an existing rank-based KBQA method, we transform the problem into a multiple-choice question format, and then construct prompts for LLM to make choices via ICL, as shown in Figure \ref{fig:method}. The upper left part of the figure shows a rank-based KBQA method. Using a ranker to score logical forms from a pool of candidate logical forms which are enumerated by parsing questions and searching over KB, it selects the logical form with the highest score and fetches the answer. Based on scores from the ranker, we select a small set of candidate logical forms, fetch corresponding answers and convert the original question into a multiple-choice question form. Next, we randomly sample several examples and construct the ICL prompt input. After LLM inference, we match the option letter at the end of the generated text, using the logical form of this option letter as the result.

However for complex questions with constraints, the performance of ICL still needs to be improved. Therefore, we add question explanations with chain-of-thought (CoT) to ICL prompt, identifying constraint information in questions. we expect LLM can choose the correct answer with the aid of constraint information obtained by CoT.\par

Through analyzing the results of preliminary experiments on 200 questions, we find that the overall performance of LLM is inferior to rank-based methods. However, there is still a considerable proportion of questions with higher accuracy than the rank-based methods. If the results of LLM can be effectively combined with existing KBQA methods, the performance will improve predictably. To complement the advantages of rank-based methods and LLM results, we adopt a result fusion strategy: evaluating the degree of certainty for the ranker in candidate ranking and using the LLM result as a substitute for questions with low certainty.\par

In summary, our contributions include:
\begin{itemize}

    \item We propose a KBQA framework with LLM via ICL to answer multiple-choice questions, which are transformed from the original question using an existing rank-based KBQA method. In addition, We use chain-of-thought to get question explanations for ICL examples, which achieve further improvement.

    \item We adopt a simple but effective result fusion strategy to complement the advantages of existing method and LLM results.
    
    \item Experiments on WebQSP and GrailQA datasets demonstrate the effectiveness of our framework, especially under the few-shot setting.
\end{itemize}

\section{Related Work}
\subsection{Knowledge base question answering}

Most state-of-the-art KBQA methods are based on semantic parsing \citep{chen-etal-2021-retrack,ye2021rng,gu-su-2022-arcaneqa}. Specifically, they enumerate candidate logical forms based on the entity in the question and then apply a ranker to score every candidate, choosing one logical form with the highest score to find the answer. we refer to them as rank-based methods here. However, sufficient training data is necessary for rank-based methods to achieve competitive performance. \citep{li2023few} first apply ICL for KBQA task in few-shot settings. It generates logical forms drafts with LLM, and then binds entities and schema items to KB iteratively until an executable one can be found.  

\subsection{In-Context Learning}

In-context learning (ICL) with LLMs \citep{brown2020language} is about applying LLM to new tasks without updating the parameters, only providing a few demonstrations of input-output pairs at inference time.
It has been found to be competitive in a broad range of tasks including information extraction \citep{ma2023large}, machine translation \cite{agrawal2022context}, numerical reasoning \citep{lewkowycz2022solving} and semantic parsing \citep{shin-van-durme-2022-shot}. 

Many studies focused on prompt construction to achieve better performance. \citep{min-etal-2022-rethinking} shows the effectiveness of constructing prompts using an input-label pairing format, and \citep{liu-etal-2022-makes} experiment with the number of examples provided, as well the idea of retrieving relevant examples to a test input to construct the prompt with. \citep{lampinen-etal-2022-language} suggests that incorporating explanatory task instructions in context can improve performance. 

\section{KBQA with In-Context Learning}

\subsection{Preliminaries and Ranker}

A knowledge base (KB) consists of a set of entities \textit{E} and relations \textit{R}. Knowledge can be represented as a set of fact triples (\textit{s}, \textit{p}, \textit{o}), where \textit{s} and \textit{o} are entities from \textit{E} and \textit{p} is a relation from \textit{R}. For question \textit{q}, the goal of KBQA is to return a set of entities \textit{E$_A$} $\subset$ \textit{E} as the answer to \textit{q}.\par

We select a wild used rank-based method to provide candidates for LLM, consisting of two steps.\newline
\paragraph{Enumerate Candidate} 
Given entities detected in the question, we query the knowledge base starting from every entity for paths reachable within two hops. we save the paths into logical forms, which constitute a set of logical form candidates \textit{C} = $\{c_i\}_{i=1}^n$.\newline 

\paragraph{Ranker} 
Following the setting in \citet{ye2021rng}, We train a BERT-based ranker to score every logical form candidate.  Given the question \textit{q} and a logical form candidate \textit{c$_i$} $\in$ \textit{C}, we concatenate \textit{q} and \textit{c$_i$}  as the input of a BERT-based encoder, taking the output logit as the similarity between them:\newline
\begin{equation}
    s(q,c_i) = \mathrm{Liner}(\mathrm{BERT}([q;c_i]))
    \label{eq:ranker_score}
\end{equation}
where BERT denotes [CLS] representation of input; Liner is a linear layer reducing representation to similarity score. We randomly sample negative logical form candidates during training without using the bootstrap strategy.\par
We select the logical form candidate with the highest score as ranker result \textit{lf$_{ranker}$}:
\begin{equation}
    lf_{ranker} = \mathop{\arg\max}_{c_i \in C} s(q,c_i)
\end{equation}

\subsection{Make Choice by LLM via ICL}
We reformulate question \textit{q} into the form of multiple-choice question \textit{q$_{choice}$}. Using LLM via in-context learning to solve \textit{q$_{choice}$} by selecting one option \textit{A$_{opt}$} from given options, we can obtain the answer to \textit{q} by returning the answer corresponding to \textit{A$_{opt}$}.

\subsubsection{Reduce the number of candidates}
Due to the huge number of candidates (up to thousands), it is impossible to use every candidate for building multiple-choice questions \textit{q$_{choice}$}. Hence We use Ranker to score every candidate \textit{c$_i$} $\in$ \textit{C}, select top \textit{k} logical form candidate for the next step. We mark this smaller candidate set as \textit{C$_k$}.

\subsubsection{Prompt Construction}
\begin{figure}[t]
    \centering
    \includegraphics[width=7cm]{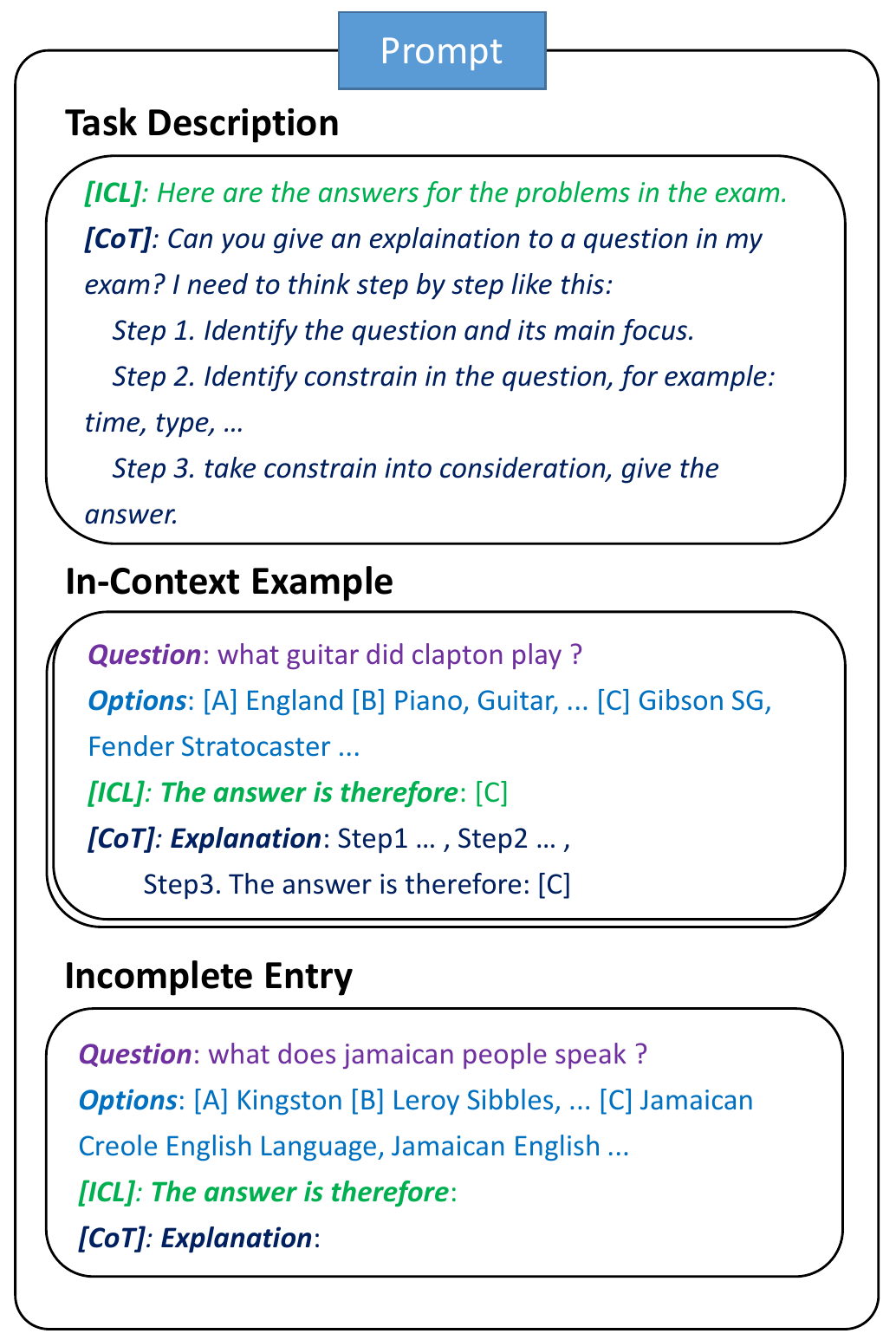}
    \caption{An example of prompt. Consist of three parts: task description, in-context example, and incomplete entry. We have two forms of prompt, ICL and CoT. The different parts between them are marked with two prefixes \textbf{\textit{[ICL]}} and \textbf{\textit{[CoT]}}.}
    \label{fig:prompt example}
\end{figure}

Based on question \textit{q} and candidates \textit{C$_k$}, we formulate multiple-choice question \textit{q$_{choice}$} and construct prompts for LLM. The prompt has three parts: \textbf{Task Description}, \textbf{In-Context Example}, and \textbf{Incomplete Entry}. Figure \ref{fig:prompt example} is an example of prompt.\par

\paragraph{Task Description} is a short description of the task. We draft a simple version without making too many attempts.\par

\paragraph{In-Context Example} consists of a question, options, and answer. We select question \textit{q} for in-context examples by random sampling from the development set and make sure the gold logic form \textit{lf$_g$} is in candidates set \textit{C$_k$}. Based on \textit{C$_k$}, we build options for \textit{q}. For \textit{lf$_i$} $\in$ \textit{C$_k$}, we shift \textit{lf$_i$} into query and get the corresponding entities set \textit{E$_i$} by execute the query in KB. We control the size of \textit{E$_i$} smaller than 5 for cost consideration. \textit{E$_i$} is consist of entity IDs like "\textit{m.01428y}". To make good use of LLMs, we fetch the surface name set for \textit{e} $\in$ \textit{E$_i$} as \textit{N$_i$} like "\textit{Jamaican English}" rather than IDs. We build the option context of \textit{lf$_i$} by joining \textit{n} $\in$ \textit{N$_i$} with comma and attaching it to an option letter \textit{opt$_l$} like "\textit{[A]}". Combining the question \textit{q} and all options, we get the multiple-choice question \textit{q$_{choice}$}. Finally, we use the option letter corresponding to \textit{lf$_g$} as the answer of \textit{q$_{choice}$}.
\par

\paragraph{Incomplete Entry}The composition is similar to the in-context example. The questions here are those need to be answered. Options are built out of entity names fetched from KB based on \textit{lf$_i$} $\in$ \textit{C$_k$}, similar to the in-context example. As the answer part, we leave it blank for LLM to complete, like "\textit{the answer is therefore}". 

\subsubsection{Question Explanation with Chain of Thought}

Inspired by previous work \citep{zhang2022automatic} with question explanation, we propose a new form of prompt: use LLM to generate question explanation with CoT to analyze the question and choose one option letter in the end. As shown in Figure \ref{fig:prompt example} (marked with prefix \textbf{\textit{[CoT]}}), we adopt a new task description to generate an explanation for the question, following 3 steps as a guide: Step 1. identify the main focus of the question; Step 2. identify constrain in question; Step 3. give the selected answer. For the explanation to question in in-context example, we obtain it by LLM via zero-shot CoT, based on the task definition above. As to incomplete entry, it ends like "\textit{explanation for problem}" instead for LLM to complete. 

\subsubsection{Inference}
We feed the prompt input \textit{prompt$_q$} to LLM and get the output text \textit{T}. For the convenience of matching results, we set a stop flag "]" where LLM will stop generating further tokens. This makes \textit{T} ends like "\textit{the answer is therefore [\textbf{A}}".
We match an option letter \textit{A$_{opt}$} for the multiple-choice question at the end of \textit{T}. We obtain the logical form corresponding to \textit{A$_{opt}$} as the LLM result \textit{lf$_{llm}$}.\par

\begin{equation}
    T = \mathrm{LLM}(prompt_{q})
\end{equation}
\begin{equation}
    lf_{llm} = \mathrm{Match}(T)
\end{equation}

For a small part of questions, LLM fail to give an available \textit{A$_{opt}$}, we use \textit{lf$_{ranker}$} as a substitute.

\subsection{Result Fusion}
Inspired by \citep{ma2023large}, we fuse the ranker and LLM results. Given question \textit{q} and its candidates \textit{C}, we use ranker to score \textit{c$_i$} $\in$ \textit{C} as shown in (\ref{eq:ranker_score}) and adopt the maximum score among all candidates as confidence score \textit{s(q)} of \textit{q}. We set a threshold $\lambda$ to \textit{s(q)}. For questions with \textit{s(q)} lower than $\lambda$, we use LLM results, otherwise using Ranker results.\par

\begin{equation}
    s(q) = \max_{c_i \in C}s(q,c_i)
\end{equation}

\begin{equation}
lf_{fuse} = \left\{
\begin{array}{lc}
lf_{llm} & s(q) < \lambda \\
lf_{ranker} & else
\end{array}
\right.
\end{equation}

\section{Experiment}

\subsection{Datasets}
We use two datasets with annotated logical forms to evaluate our method.\par
\textbf{WebQuestionsSP (WebQSP)} \citep{yih2016value} is a widely used KBQA dataset with 4,727 questions, officially divided into train/test (3098/1639) set. We randomly sample 200 questions from the train set as local development set.\par
\textbf{GrailQA} \citep{gu2021beyond} is a large-scale KBQA dataset with 64,331 questions, which is created to evaluate three levels of generalization in KBQA: \textit{i.i.d.}, \textit{compositional}, and \textit{zero-shot}. It contains 44k, 6k, and 13k for training, development, and testing, respectively. The official test of GrailQA is a closed set, thus we split the official development set equally as the local development set and test set.\par

\subsection{Experiment Settings}
\paragraph{Metrics} Consistent with previous work, we use \textbf{F1} as the evaluation metric on WebQSP. On GrailQA, we use official metrics Exact Match (\textbf{EM}) and F1-score (\textbf{F1}). 

\paragraph{Few-Shot Setting} We train Ranker with different percentages of the train set, selecting the checkpoint with the best \textbf{F1} on local development set and evaluating on local test set. The percentage and size of the few-shot train set on two datasets are shown in Table \ref{tab:few_shot}. For few-shot settings on GrailQA, we only report overall F1 \& EM.

\begin{table}[]
    \centering
    \begin{tabular}{cc|cc}
        \hline
        \multicolumn{2}{c|}{WebQSP} & \multicolumn{2}{c}{GrailQA}\\\hline
        per & size & per & size \\\hline
        5\% & 144 & 0.5\% & 221\\
        10\% & 289 & 1\% & 443\\
        30\% & 869 & 5\% & 2216\\
        50\% & 1449 & 10\% & 4433\\\hline
    \end{tabular}
    \caption{Percentage and size of few-shot train sets on WebQSP and GrailQA, arranged by size from small to large. The percentage here is an approximate value. }
    \label{tab:few_shot}
\end{table}

\paragraph{Method} We report the result of the following methods.\par
\begin{itemize}
    \item \textbf{rank}: The result of the rank-based method, marked it as rank for short. We use Ranker to score every candidate and select the top 1 candidate to fetch the answer.
    \item \textbf{ICL}: Let LLM answer multiple-choice questions with the base prompt.
    \item \textbf{CoT}: Similar to \textbf{ICL}, but use prompt in the form of question explain with CoT.
    \item \textbf{w/ fuse}: do result fusion between \textbf{rank} and LLM (\textbf{ICL}, \textbf{CoT}).
\end{itemize}

\paragraph{Implementation Details} We have 4 options for multiple-choice questions based on candidates provided by the rank-based method. For prompt construction, We randomly sample 2 exemplary questions from the development sets of WebQSP and GrailQA respectively to build in-context examples. In the inference step, we leverage ChatGPT (gpt-3.5-turbo-0301) from OpenAI API as our LLM with the temperature setting to 0. In the result fusion step, we set $\lambda$ based on the proportion of problems involved in the fusion which is set to 5\%, in order to control the scale of result fusion. Specifically, if the overall performance of rank/LLM is better, we use the result of rank for 95\%/5\% questions with higher confidence scores and use the result of LLM for the remaining questions. For few-shot settings on two datasets, we report average results on 5 random seeds.

\subsection{Main Result}

\begin{table}[]
\centering
\begin{tabular}{lccccc}
    \hline\centering
    
    Methods & {\textbf{5\%}} & {\textbf{10\%}} & {\textbf{30\%}} & {\textbf{50\%}} & {\textbf{100\%}} \\
    \hline\centering
    rank & 48.73 & 56.09 & 63.95 & 67.49 & 69.61 \\\hline
    ICL    & 58.56 & 61.11 & 63.17 & 63.85 & 64.83\\
    \hspace{4pt}w/ fuse & \underline{58.75} & \underline{61.38} & \underline{64.62} & \underline{68.43} & \textbf{70.42} \\\hline
    CoT & 59.78 & 61.90 & 64.09 & 65.53 & 66.25 \\
    \hspace{4pt}w/ fuse & \textbf{59.79} & \textbf{62.08} & \textbf{65.00} & \textbf{68.45} & \underline{70.32}\\
\hline
\end{tabular}
\caption{F1 scores on WebQSP}
\label{tab:main_res_webqsp}
\end{table}

\begin{table*}[]
\centering
\begin{tabular}{lcccccccccc}
    \hline\centering
    
    Methods & \multicolumn{2}{c}{\textbf{0.5\%}}& \multicolumn{2}{c}{\textbf{1\%}} & \multicolumn{2}{c}{\textbf{5\%}} & \multicolumn{2}{c}{\textbf{10\%}}  & \multicolumn{2}{c}{\textbf{100\%}}\\

    & \textbf{EM} & \textbf{F1} & \textbf{EM} & \textbf{F1} & \textbf{EM} & \textbf{F1} & \textbf{EM} & \textbf{F1} & \textbf{EM} & \textbf{F1}\\
    \hline\centering
    rank & 44.71 & 49.96 & 50.60 & 56.17 & 60.70 & 66.26 & 62.50 & 67.51 & 62.92 &  67.78\\\hline
    ICL    & 41.19 & 47.84 & 41.51 & 47.97 & 42.02 & 48.87 & 43.35 & 49.52 & 43.14 & 49.37\\
    \hspace{4pt}w/ fuse & \underline{49.08} & \underline{54.76} & \underline{53.28} & \underline{59.59} & \underline{61.62} & \underline{67.95} & \underline{63.57} & \underline{69.18} & \textbf{64.70} & \textbf{69.97}\\\hline
    CoT & 44.26 & 50.71 & 44.91 & 51.48 & 45.12 & 52.03 & 46.39 & 52.48 & 46.04 &   52.18\\
    \hspace{4pt}w/ fuse & \textbf{49.20} & \textbf{54.82} & \textbf{53.37} & \textbf{59.62} & \textbf{61.68} & \textbf{68.05} & \textbf{63.69} & \textbf{69.32} & \underline{64.64} & \underline{69.93}\\
    \hline
\end{tabular}
\caption{Overall EM and F1 on GrailQA (local dev).}

\label{tab:main_res_grailqa}
\end{table*}

\begin{table*}[]
    \centering
    \begin{tabular}{lcccccccc}
    \hline
     & \multicolumn{2}{c}{\textbf{Overall}} & \multicolumn{2}{c}{\textbf{I.I.D.}} & \multicolumn{2}{c}{\textbf{Compositional}} & \multicolumn{2}{c}{\textbf{Zero-Shot}}\\
     & \textbf{EM} & \textbf{F1} & \textbf{EM} & \textbf{F1} & \textbf{EM} & \textbf{F1} & \textbf{EM} & \textbf{F1}\\\hline
    rank & 62.92 & 67.78 & \textbf{71.27} & 74.57 & 56.27 & 60.45 & 62.04 & 67.85\\\hline
    ICL & 43.14 & 49.37 & 48.43 & 53.38 & 37.65 & 42.89 & 43.11 & 50.31\\
    \hspace{4pt}w/fuse & \textbf{64.70} & \textbf{69.97} & 71.02 & 74.84 & 57.99 & 62.29 & \textbf{64.72} & \textbf{71.02}\\\hline
    CoT & 46.04 & 52.18 & 50.06 & 55.22 & 39.63 & 44.87 & 46.94 & 53.88\\
    \hspace{4pt}w/fuse & 64.64 & 69.93 & 70.89 & \textbf{74.86} & \textbf{58.12} & \textbf{62.39} & 64.61 & 70.90\\\hline
    \end{tabular}
    \caption{Results of three levels on GrailQA (local dev) with the full train set (100\%).}
    \label{tab:main_res_grailqa_full}
\end{table*}

\noindent Result on WebQSP are shown in Table \ref{tab:main_res_webqsp}. In two few-shot settings (5\%, 10\%), ICL can significantly superior to Ranker. Among them, ICL can surpass Ranker by 9.83 with 5\% training data. However, when the training data is more sufficient (30\%, 50\%, 100\%), ICL will be inferior to Ranker. Moreover, as the amount of training data increases, the gap between ICL and Ranker will become larger, increasing from 0.79 (30\%) to 4.78(100\%).\par
As for ICL w/ fuse (fusing Ranker and ICL), the result can achieve stable improvement, both for Ranker and ICL. In the three settings (30\%, 50\%, 100\%) where ICL is inferior to Ranker, ICL w/fuse can surpass Ranker with an average improvement of 0.80. It is worth noting that we only select 5\% of the total questions using ICL results in these three groups. \par
CoT exceeds ICL in all settings, with an average improvement of 1.21, and the overall trend is similar to ICL. It is inferior to rank in settings where the training data is sufficient (50\%, 100\%). After result fusion (CoT w/ fuse), it can also exceed rank. The advantage of CoT w/fuse over ICL w/fuse is more obvious at lower ratios of training set, and the difference is smaller when the training data is sufficient.\par

Overall results on GrailQA are shown in Table \ref{tab:main_res_grailqa}. Performance is consistent with our analysis on WebQSP. In order to evaluate the generalization ability, we report the results of three levels on the full train set (100\%) of GrailQA in Table \ref{tab:main_res_grailqa_full}. Our method outperforms rank by 1.85 EM/1.94 F1 in \textit{Compositional} level, and 2.68 EM/3.17 F1 in \textit{Zero-Shot} level respectively. It is also competitive on \textit{I.I.D.} level, slightly inferior to rank in EM but superior to it in F1. The above results prove that our method has a strong generalization ability and keeps competitive on I.I.D. questions at the same time.

\subsection{Analysis}
\begin{figure}[t]
    \centering
    \includegraphics[width=7cm]{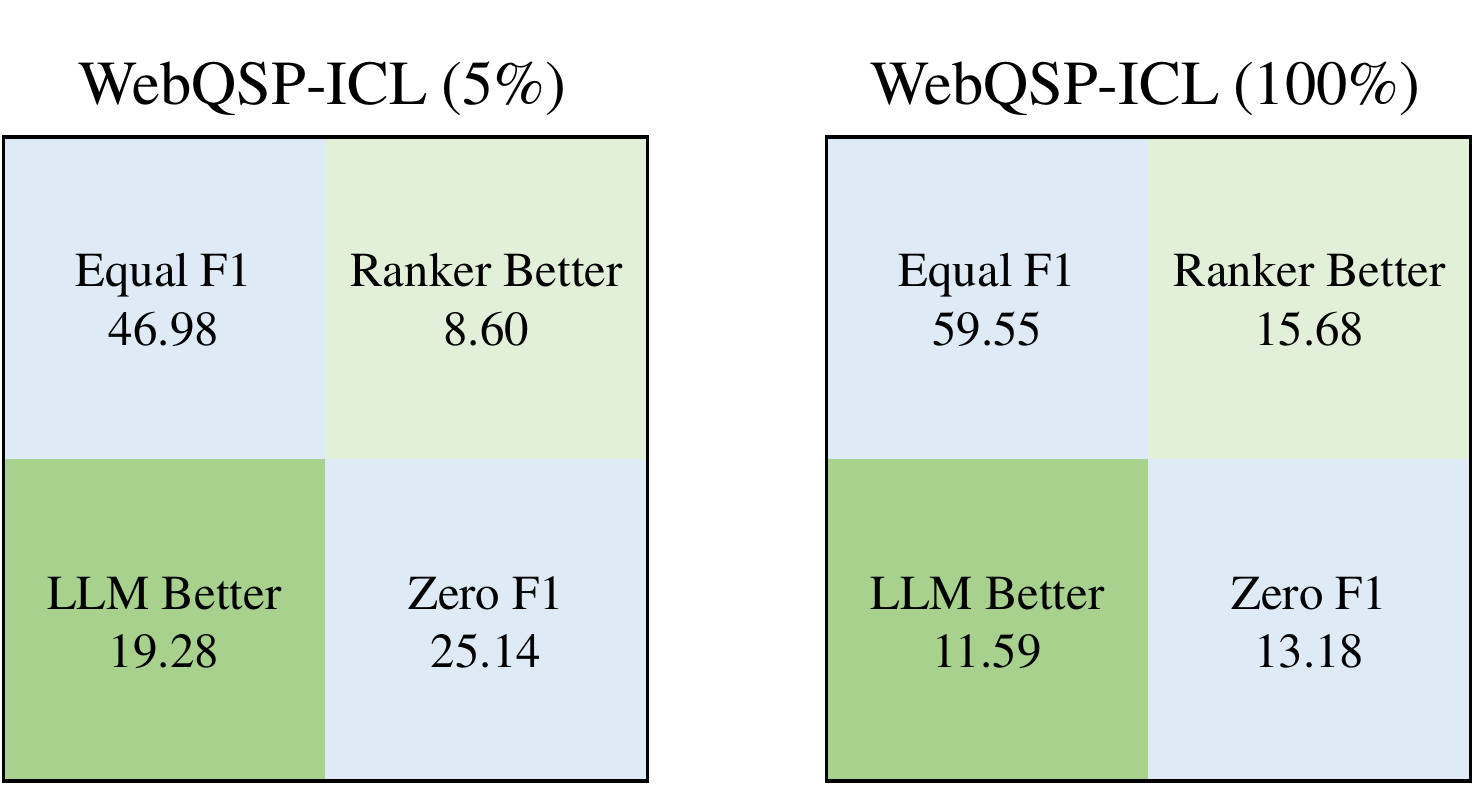}
    \caption{Comparison between rank and LLM (ICL) result on WebQSP dataset with the percentage of four output situations. In few-shot setting (5\%), LLM mostly improves or keeps the result, occasionally introducing errors.}
    \label{fig:ranker_llm_compare}
\end{figure}
\begin{table}[t]
    \centering
    \begin{tabular}{ccc}\hline
    settings & H & L\\\hline
    5\% & 19.89\% & 7.99\% \\
    100\% & 17.69\% & 9.58\% \\\hline
    \end{tabular}
    \caption{Statistics of result fusion on the test set of WebQSP dataset (ICL w/fuse), reporting proportion(\%) of: \textbf{H}: the fused result is the one with higher F1 among rank and LLM. \textbf{L}: use the result with lower F1.}
    \label{tab:res_fuse}
\end{table}

\begin{table*}[th]
    \centering
    \begin{tabular}{|l|}
    \hline
    \textbf{Examples of CoT Output} \\\hline
    (a) \textbf{Question}: What team does jeremy lin play for 2013? \\
    \textbf{Options}: \small\textbf{[A] Houston Rockets} [B] Vive Targi Kielce [C] New York Knicks, Houston Rockets [D] 762195 \\
    \textbf{Output}: \small Step 1: The main focus of the question is to identify the \textbf{team that Jeremy Lin played for in 2013}.\\
    \small Step 2: The question specifies the \textcolor{red}{\textbf{year 2013}} as a constraint. \\
    \small Step 3: The answer is therefore [A \\
    \hline
    (b) \textbf{Question}: Where george lopez was born?  \\
    \textbf{Options}: \small[A] San Fernando High School [B] 1961-04-23 [C] Mission Hills, Los Angeles \textbf{[D] Mission Hills} \\
    \textbf{Output}: \small Step 1: The main focus of the question is to identify the birthplace of George Lopez.\\
    \small Step 2: The question specifically asks for a \textcolor{red}{\textbf{location}}. \\
    \small Step 3: The answer is therefore [D \\
    \hline

    (c) \textbf{Question}: Where are boeing headquarters?\\
    \textbf{Options}: \small [A] Seattle [B] United States of America \textbf{[C] Chicago} [D] King County\\
    \textbf{Output}: \small Step 1: The main focus of the question is to identify the location of Boeing headquarters.\\
        \small Step 2: The question specifically asks for a \textcolor{red}{\textbf{location}}.\\
        \small Step 3: The answer is therefore [A\\\hline
    (d) \textbf{Question}: Where did kansas city royals come from?\\
    \textbf{Options}: \small [A] Kauffman Stadium \textbf{[B] Kansas City} [C] 1969 [D] Ewing Kauffman\\
    \textbf{Output}: \small Step 1: The main focus of the question is to identify the origin of the Kansas City Royals. \\
        \small Step 2: The question specifically asks for \textcolor{red}{\textbf{a location and a time period}}. \\
        \small Step 3: The answer is therefore [C\\\hline
    
    \end{tabular}
    \caption{Case Study: Examples of LLM output with CoT. The best option is marked in \textbf{bold}. CoT is able to provide information such as constraint (a) or answer type (b) to help with selection. However, the provided information can not help selection if all options match it (c). Also, CoT may introduce wrong information sometimes (d).}

    \label{tab:CoT_case}
\end{table*}

\subsubsection{Comparing Outputs of Ranker and LLMs}

We carefully compare the performance between rank and LLM outputs (ICL version) on two sets of WebQSP dataset (5\% and 100\%). Figure \ref{fig:ranker_llm_compare} shows the percentage of four types of output situations on test set of WebQSP:
\begin{itemize}
    \item top left: F1 of rank and LLM outputs are equal but not 0,
    \item top right: rank is better,
    \item bottom left: LLM is better,
    \item bottom right: Their F1 are both 0.
\end{itemize}
Most of the time, LLM outputs are consistent with those of rank. In few-shot setting, LLM is better than rank on 19.28\% of questions. It is worth noting that, although F1 of LLM output is overall inferior to rank in full setting (100\%), it still performs better than rank on 11.59\% questions. It is similar to rank output under few-shot settings, where the outputs of 8.60\% questions are better than LLM. This indicates that combining the output of rank and LLM efficiently can improve performance. This indicates that our result fusion strategy can complement the advantages of rank and LLM, and select answers with higher performance between the two in more questions, achieving overall performance improvement.

\subsubsection{Effectiveness of Result Fusion}

The experiment results in Table \ref{tab:main_res_webqsp} and Table \ref{tab:main_res_grailqa} demonstrate that there is a stable improvement in result fusion under every group setting. We select WebQSP 5\%/100\% as an example of few-shot/full setting to analyze the effect of result fusion in more detail. The specific settings are consistent with Table \ref{tab:main_res_webqsp}. As shown in Table \ref{tab:res_fuse}, we statistics the proportion of questions in fused results using \textbf{H}igher/\textbf{L}ower performance results among rank and LLM. At the 5\% setting, overall F1 of LLM is higher than rank, having 19.28\% H question ("LLM better" in the left part of Figure \ref{fig:ranker_llm_compare}). After result fusion, 0.61\%  (19.89\% -19.28\%) of H questions are increased. As to full setting, result fusion brings an increase of 2.01\% (17.69\% - 15.68\%) H questions. This indicates that our result fusion strategy can complement the advantages of rank and LLM, selecting answers with higher performance between the two in more questions, and achieving overall performance improvement.

\subsubsection{Does CoT Make a Different?}

We use two forms of prompts (ICL and CoT) for LLM to do multiple-choice questions. It can be seen in the main result (Table \ref{tab:main_res_webqsp} and Table \ref{tab:main_res_grailqa}) that CoT shows stable improvements compared to ICL. To clarify the strengths and weaknesses of CoT, We conduct a case study on the output of LLM using CoT prompt (Table \ref{tab:CoT_case}). In example (a), the question explanation generated by CoT can mention "\textit{2013}" as a time constraint to the question. CoT can also identify the answer type to help with selection: "\textit{location}" is the answer type provided by CoT in (b), which can help to exclude option [B]. \par
However, the information provided by CoT may not help selection in some cases. As shown in example (c), CoT provides the answer type "\textit{location}", but each option is related to a location. It may also fail when CoT introduces wrong information such as "\textit{time period}" in example (d), which leads to an incorrect selection of time-related option [C].\par

\subsubsection{Match Option from LLM Output}

Given LLM output, we automatically match an option letter \textit{A$_{opt}$} from the end of output using a regular expression, and obtain the corresponding logical form as the LLM result based on \textit{A$_{opt}$}. We analyze states of our automatic option letter matching results on the test set of WebQSP dataset, as shown in Table \ref{tab:auto_match_states}. For ICL and CoT prompt inputs, we can match valid options (\textbf{Match}) with proportions of 96.36\% and 93.52\% respectively. This indicates that our matching method can extract valid results of LLM in most cases. 
For questions that failed to match the options (\textbf{Fail}), we check the \textbf{Fail} cases of ICL and CoT in the full train setting. Every case of ICL (a) regards the correct answer is not being provided, and giving alternative answers sometimes. As for cases of CoT can be divided into four types: type (a) which is introduced above (40.90\%); (b) return more than one option letter (16.67\%); (c) give option context rather than option letter (16.67\%).

It is worth noting that there are a few questions that match an option letter not provided in the options list (\textbf{OOL}), with 0.14\% in ICL and 0.06\% in CoT. Due to the limited number of such cases, we can analyze each of them. With the aid of the question explanation generated by CoT, we believe it is another expression of LLM that all options provided are incorrect.

\begin{table}[]
\centering
\begin{tabular}{lccc}
\hline
\textbf{} & {\textbf{Match}} & {\textbf{Fail}} & {\textbf{OOL}} \\
\hline
\textbf{ICL} & 96.36 & 3.50 & 0.14 \\
\textbf{CoT} & 93.52 & 6.43 & 0.06 \\
\hline
\end{tabular}
\caption{proportion(\%) of different states of automatic option letter matching result from LLM outputs on test set of WebQSP, reporting average results of different settings. \textbf{ICL} and \textbf{CoT}: LLM output of two prompt input forms. \textbf{Match}: match a valid option letter, \textbf{Fail}: fail to match an option letter, \textbf{OOL}: match an option letter but not provided in the options list.}
\label{tab:auto_match_states}
\end{table}

\section{Conclusion}
In this paper, we propose McL-KBQA framework for knowledge base question answering. We transform the original question into a multiple-choice question with a rank-based existing KBQA method and use LLM via ICL to answer the question by making a choice. In addition, we use chain-of-thought to get question explanations for example in ICL. Finally, we adopt a simple but effective result fusion strategy to complement the advantages of the rank-based method and LLM results. The experimental results on two datasets, WebQSP and GrailQA, suggest the effectiveness of our framework, especially under the few-shot setting.

\bibliography{anthology,custom}

\begin{thebibliography}{18}
\expandafter\ifx\csname natexlab\endcsname\relax\def\natexlab#1{#1}\fi

\bibitem[{Agrawal et~al.(2022)Agrawal, Zhou, Lewis, Zettlemoyer, and
  Ghazvininejad}]{agrawal2022context}
Sweta Agrawal, Chunting Zhou, Mike Lewis, Luke Zettlemoyer, and Marjan
  Ghazvininejad. 2022.
\newblock In-context examples selection for machine translation.
\newblock \emph{arXiv preprint arXiv:2212.02437}.

\bibitem[{Brown et~al.(2020)Brown, Mann, Ryder, Subbiah, Kaplan, Dhariwal,
  Neelakantan, Shyam, Sastry, Askell et~al.}]{brown2020language}
Tom Brown, Benjamin Mann, Nick Ryder, Melanie Subbiah, Jared~D Kaplan, Prafulla
  Dhariwal, Arvind Neelakantan, Pranav Shyam, Girish Sastry, Amanda Askell,
  et~al. 2020.
\newblock Language models are few-shot learners.
\newblock \emph{Advances in neural information processing systems},
  33:1877--1901.

\bibitem[{Chen et~al.(2021)Chen, Liu, Yu, Lin, Lou, and
  Jiang}]{chen-etal-2021-retrack}
Shuang Chen, Qian Liu, Zhiwei Yu, Chin-Yew Lin, Jian-Guang Lou, and Feng Jiang.
  2021.
\newblock \href {https://doi.org/10.18653/v1/2021.acl-demo.39} {{R}e{T}ra{C}k:
  A flexible and efficient framework for knowledge base question answering}.
\newblock In \emph{Proceedings of the 59th Annual Meeting of the Association
  for Computational Linguistics and the 11th International Joint Conference on
  Natural Language Processing: System Demonstrations}, pages 325--336, Online.
  Association for Computational Linguistics.

\bibitem[{Gu et~al.(2021)Gu, Kase, Vanni, Sadler, Liang, Yan, and
  Su}]{gu2021beyond}
Yu~Gu, Sue Kase, Michelle Vanni, Brian Sadler, Percy Liang, Xifeng Yan, and
  Yu~Su. 2021.
\newblock Beyond iid: three levels of generalization for question answering on
  knowledge bases.
\newblock In \emph{Proceedings of the Web Conference 2021}, pages 3477--3488.

\bibitem[{Gu and Su(2022)}]{gu-su-2022-arcaneqa}
Yu~Gu and Yu~Su. 2022.
\newblock \href {https://aclanthology.org/2022.coling-1.148} {{A}rcane{QA}:
  Dynamic program induction and contextualized encoding for knowledge base
  question answering}.
\newblock In \emph{Proceedings of the 29th International Conference on
  Computational Linguistics}, pages 1718--1731, Gyeongju, Republic of Korea.
  International Committee on Computational Linguistics.

\bibitem[{Ji et~al.(2023)Ji, Lee, Frieske, Yu, Su, Xu, Ishii, Bang, Madotto,
  and Fung}]{ji2023survey}
Ziwei Ji, Nayeon Lee, Rita Frieske, Tiezheng Yu, Dan Su, Yan Xu, Etsuko Ishii,
  Ye~Jin Bang, Andrea Madotto, and Pascale Fung. 2023.
\newblock Survey of hallucination in natural language generation.
\newblock \emph{ACM Computing Surveys}, 55(12):1--38.

\bibitem[{Lampinen et~al.(2022)Lampinen, Dasgupta, Chan, Mathewson, Tessler,
  Creswell, McClelland, Wang, and Hill}]{lampinen-etal-2022-language}
Andrew Lampinen, Ishita Dasgupta, Stephanie Chan, Kory Mathewson, Mh~Tessler,
  Antonia Creswell, James McClelland, Jane Wang, and Felix Hill. 2022.
\newblock \href {https://aclanthology.org/2022.findings-emnlp.38} {Can language
  models learn from explanations in context?}
\newblock In \emph{Findings of the Association for Computational Linguistics:
  EMNLP 2022}, pages 537--563, Abu Dhabi, United Arab Emirates. Association for
  Computational Linguistics.

\bibitem[{Lewkowycz et~al.(2022)Lewkowycz, Andreassen, Dohan, Dyer,
  Michalewski, Ramasesh, Slone, Anil, Schlag, Gutman-Solo
  et~al.}]{lewkowycz2022solving}
Aitor Lewkowycz, Anders Andreassen, David Dohan, Ethan Dyer, Henryk
  Michalewski, Vinay Ramasesh, Ambrose Slone, Cem Anil, Imanol Schlag, Theo
  Gutman-Solo, et~al. 2022.
\newblock Solving quantitative reasoning problems with language models.
\newblock \emph{arXiv preprint arXiv:2206.14858}.

\bibitem[{Li et~al.(2023)Li, Ma, Zhuang, Gu, Su, and Chen}]{li2023few}
Tianle Li, Xueguang Ma, Alex Zhuang, Yu~Gu, Yu~Su, and Wenhu Chen. 2023.
\newblock Few-shot in-context learning for knowledge base question answering.
\newblock \emph{arXiv preprint arXiv:2305.01750}.

\bibitem[{Liu et~al.(2022)Liu, Shen, Zhang, Dolan, Carin, and
  Chen}]{liu-etal-2022-makes}
Jiachang Liu, Dinghan Shen, Yizhe Zhang, Bill Dolan, Lawrence Carin, and Weizhu
  Chen. 2022.
\newblock \href {https://doi.org/10.18653/v1/2022.deelio-1.10} {What makes good
  in-context examples for {GPT}-3?}
\newblock In \emph{Proceedings of Deep Learning Inside Out (DeeLIO 2022): The
  3rd Workshop on Knowledge Extraction and Integration for Deep Learning
  Architectures}, pages 100--114, Dublin, Ireland and Online. Association for
  Computational Linguistics.

\bibitem[{Ma et~al.(2023)Ma, Cao, Hong, and Sun}]{ma2023large}
Yubo Ma, Yixin Cao, YongChing Hong, and Aixin Sun. 2023.
\newblock Large language model is not a good few-shot information extractor,
  but a good reranker for hard samples!
\newblock \emph{arXiv preprint arXiv:2303.08559}.

\bibitem[{Min et~al.(2022)Min, Lyu, Holtzman, Artetxe, Lewis, Hajishirzi, and
  Zettlemoyer}]{min-etal-2022-rethinking}
Sewon Min, Xinxi Lyu, Ari Holtzman, Mikel Artetxe, Mike Lewis, Hannaneh
  Hajishirzi, and Luke Zettlemoyer. 2022.
\newblock \href {https://aclanthology.org/2022.emnlp-main.759} {Rethinking the
  role of demonstrations: What makes in-context learning work?}
\newblock In \emph{Proceedings of the 2022 Conference on Empirical Methods in
  Natural Language Processing}, pages 11048--11064, Abu Dhabi, United Arab
  Emirates. Association for Computational Linguistics.

\bibitem[{Omar et~al.(2023)Omar, Mangukiya, Kalnis, and
  Mansour}]{omar2023chatgpt}
Reham Omar, Omij Mangukiya, Panos Kalnis, and Essam Mansour. 2023.
\newblock Chatgpt versus traditional question answering for knowledge graphs:
  Current status and future directions towards knowledge graph chatbots.
\newblock \emph{arXiv preprint arXiv:2302.06466}.

\bibitem[{Shin and Van~Durme(2022)}]{shin-van-durme-2022-shot}
Richard Shin and Benjamin Van~Durme. 2022.
\newblock \href {https://doi.org/10.18653/v1/2022.naacl-main.396} {Few-shot
  semantic parsing with language models trained on code}.
\newblock In \emph{Proceedings of the 2022 Conference of the North American
  Chapter of the Association for Computational Linguistics: Human Language
  Technologies}, pages 5417--5425, Seattle, United States. Association for
  Computational Linguistics.

\bibitem[{Tan et~al.(2023)Tan, Min, Li, Li, Hu, Chen, and
  Qi}]{tan2023evaluation}
Yiming Tan, Dehai Min, Yu~Li, Wenbo Li, Nan Hu, Yongrui Chen, and Guilin Qi.
  2023.
\newblock Evaluation of chatgpt as a question answering system for answering
  complex questions.
\newblock \emph{arXiv preprint arXiv:2303.07992}.

\bibitem[{Ye et~al.(2021)Ye, Yavuz, Hashimoto, Zhou, and Xiong}]{ye2021rng}
Xi~Ye, Semih Yavuz, Kazuma Hashimoto, Yingbo Zhou, and Caiming Xiong. 2021.
\newblock Rng-kbqa: Generation augmented iterative ranking for knowledge base
  question answering.
\newblock \emph{arXiv preprint arXiv:2109.08678}.

\bibitem[{Yih et~al.(2016)Yih, Richardson, Meek, Chang, and Suh}]{yih2016value}
Wen-tau Yih, Matthew Richardson, Christopher Meek, Ming-Wei Chang, and Jina
  Suh. 2016.
\newblock The value of semantic parse labeling for knowledge base question
  answering.
\newblock In \emph{Proceedings of the 54th Annual Meeting of the Association
  for Computational Linguistics (Volume 2: Short Papers)}, pages 201--206.

\bibitem[{Zhang et~al.(2022)Zhang, Zhang, Li, and Smola}]{zhang2022automatic}
Zhuosheng Zhang, Aston Zhang, Mu~Li, and Alex Smola. 2022.
\newblock Automatic chain of thought prompting in large language models.
\newblock \emph{arXiv preprint arXiv:2210.03493}.

\end{thebibliography}
\bibliographystyle{acl_natbib}

\end{document}